\def\year{2022}\relax
\documentclass[letterpaper]{article} % DO NOT CHANGE THIS
\usepackage{aaai22}  % DO NOT CHANGE THIS
\usepackage{times}  % DO NOT CHANGE THIS
\usepackage{helvet}  % DO NOT CHANGE THIS
\usepackage{courier}  % DO NOT CHANGE THIS
\usepackage[hyphens]{url}  % DO NOT CHANGE THIS
\usepackage{graphicx} % DO NOT CHANGE THIS
\usepackage{natbib}  % DO NOT CHANGE THIS AND DO NOT ADD ANY OPTIONS TO IT
\usepackage{caption} % DO NOT CHANGE THIS AND DO NOT ADD ANY OPTIONS TO IT
\DeclareCaptionStyle{ruled}{labelfont=normalfont,labelsep=colon,strut=off} % DO NOT CHANGE THIS
\usepackage{algorithm}
\usepackage{algorithmic}

\usepackage{newfloat}
\usepackage{listings}
\floatstyle{ruled}
\newfloat{listing}{tb}{lst}{}
\floatname{listing}{Listing}
\usepackage[utf8]{inputenc}

\title{An Interactive Explanatory AI System for Industrial Quality Control}
\author{
    Dennis Müller, Michael März, Stephan Scheele, Ute Schmid
    \\
}
\affiliations{
    Fraunhofer Institute for Integrated Circuits IIS\\
    Sensory Perception \& Analytics $\mid$ Comprehensible AI\\
    \{dennis.mueller, michael.maerz, stephan.scheele, ute.schmid\}@iis.fraunhofer.de
}
\usepackage[dvipsnames]{xcolor}
\usepackage[inline]{enumitem}
\usepackage{listings}

\lstdefinelanguage{prolog}{ % meh, color in text is discouraged
}

\begin{document}

\maketitle

\begin{abstract}
    Machine learning based image classification algorithms, such as deep neural network approaches, will be increasingly employed in critical settings such as quality control in industry, where transparency and comprehensibility of decisions are crucial.
    Therefore, we aim to extend the defect detection task towards an interactive human-in-the-loop approach that allows us to integrate rich background knowledge and the inference of complex relationships going beyond traditional purely data-driven approaches.
	We propose an approach for an interactive support system for classifications in an industrial quality control setting that combines the advantages of both (explainable) knowledge-driven and data-driven machine learning methods, in particular inductive logic programming and convolutional neural networks, with human expertise and control. The resulting system can assist domain experts with decisions, provide transparent explanations for results, and integrate feedback from users; thus reducing workload for humans while both respecting their expertise and without removing their agency or accountability.
%	\TODO{}
\end{abstract}

\section{Introduction}
%\TODO{}

Knowledge-driven machine learning methods -- in the context of this paper in particular \emph{Inductive Logic Programming} (\emph{ILP}; for an overview, see \cite{cropper2020inductive}) -- and statistical data-driven methods, particularly deep learning, have complementary advantages and disadvantages: The scope of the former is naturally limited to specific representations of the training data. ILP in particular requires the data to be represented as Horn clauses, the translation to which is non-trivial, often prohibitively difficult, and informed by prior knowledge of a user. On the other hand, ILP needs relatively little data, and the result is itself a set of Horn clauses that are easily explainable and can be verbalized in a manner comprehensible to non-experts. Furthermore, retraining an ILP system is usually relatively fast and cost-effective, making interactive updates to the system feasible to include direct user feedback. We demonstrated the applicability of an interactive ILP system which offers rich verbal explanations in the context of an assistive system for data management \cite{dare2del}.

In contrast, deep learning has a virtually unlimited scope and can act on nearly any kind of data (text, images, sound), and find and exploit highly complex patterns in an input, without a user needing to extract the assumed-to-be-important information beforehand. As a trade-off, deep learning models are effectively black boxes, offering no justifications or explanations. Moreover, they require huge amounts of data and are costly to train in the first place, making adaptation and incremental refinements to a once trained system much more difficult.

In this paper, we combine the advantages of both methods into a support system for domain experts, exemplary in the context of quality control of industrial components. 
A first proof-of-concept for the feasibility of this novel type of neural-symbolic integration has been given for classification of blocks world towers combining visual explanations generated by the LIME approach and verbal explanations generated by the ILP system Aleph \cite{limealeph}.
%\TODO{cite the lime-aleph \cite{limealeph} and dare2del \cite{dare2del} papers somewhere: DONE?!}
As a guiding use case for our experiments, we envision a setting where expensive to produce industrial components may contain various kinds of visually recognizable production flaws. Human domain experts are assumed to be able to decide -- based on difficult to formalize experiential prior knowledge -- whether the flaws in any given component are either acceptable (``ok'') or sufficiently severe for it to be discarded (``defective'').

An intended support system should then \begin{enumerate*} \item present images of such components to a user, \item detect and highlight the flaws in the image and \item provide a suggested \emph{classification} and a \emph{textual justification} thereof.\end{enumerate*} The user is then prompted to either accept the output as accurate (``right for the right reasons''), accept the classification but reject the justification (``right for the wrong reasons'') or reject both (``wrong''). The system thus suggests a comprehensible action, but importantly keeps the human in the loop and takes their feedback into account for the next image presented.

\begin{figure*}[htb]
\begin{center}
	\includegraphics{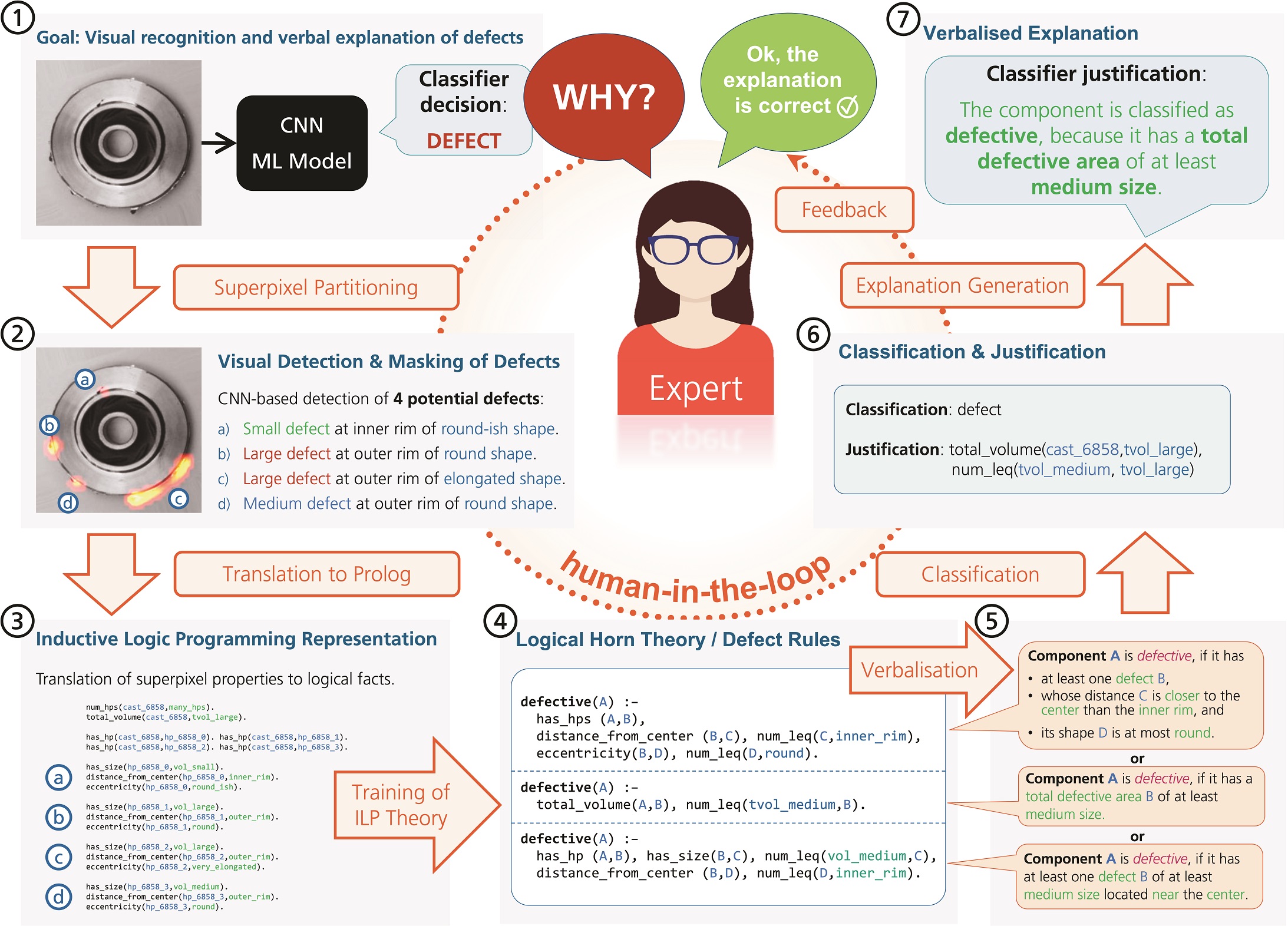}
\end{center}
	\caption{Overview of our approach and use case.}\label{img_overview}
\end{figure*}

\section{Overview}

\begin{figure}[htb]
    \centering
    \begin{tabular}{rrl}
        \includegraphics{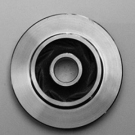}& 
        \includegraphics{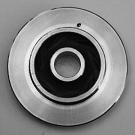}&
        \includegraphics{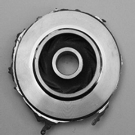}
    \end{tabular}
    \caption{Examples from \cite{castingdataset} from left to right: (i)~no defect, (ii) defect at inner rim, (iii) defect at outer rim.}
    \label{img_entry}
\end{figure}

We implement a demonstrator instance of such a system using a data set of 7348 images of cast metal industrial components \cite{castingdataset}. Figure \ref{img_entry} shows three example instances from the dataset, (i) without defects, and two with different types of defects: (ii) has a blowhole at the inner rim and (iii) shows abrasions on the outer rim. 
%with three kinds of defects: a hole on the right side, a scratch on the left inner rim and abrasions on the outer rim.

Figure \ref{img_overview} gives an overview of the approach specifically for our exemplary use case implemented by the following seven process steps:  
%The system is implemented as follows:

\begin{enumerate}
%	\item We use \cite{castingdataset}, a data set of 7348 images of cast metal industrial components, classified according to whether they show \emph{any} defects at all, as a basis (see Figure \ref{img_ex} for an example entry).
%	\item We manually create image masks highlighting the defects for 86 of the images classified as ``defective'' in the data set.
	\item First, we train a convolutional neural net (CNN) on the data set using the provided ground truth labels and 86 manually created image masks for defects, using the model from \cite{Boi2021}. The trained model in particular allows for extracting image masks for the remaining entries of the data set.
	
	After this step, we can discard the images labelled ``ok'' (no defects at all) from the data set and %, focusing on the remaining 4211 entries. 
	relabel the same 86 entries as either tolerable (``ok'') or ``defective'' as training data for the ILP component.
	\item We partition the image masks obtained from the ML model into its connected components (\emph{superpixels}, representing the individual located defects) and compute various of their properties (size, location, etc.).
	\item We represent each image as a list of Prolog clauses, by associating each type of property from the previous step with a corresponding predicate (see also Listing \ref{lst_prolog}).
	\item We train an ILP system (specifically \emph{Aleph} \cite{aleph}) on those Prolog representations of the previously relabelled images (see step 1) to obtain an initial classification hypothesis.
	\item We generate a human-readable verbalization as explanation for users. %, ideally capturing the criteria for our classification from step 4.
	\item We evalute any given image on the learned clauses from the previous step, keeping track of any contained atom's validity. We thus obtain both a classification as well as (one or multiple) Prolog-style justifications for a clause's applicability (or lack thereof). 
	\item Finally, we %generate verbalizations of 
	verbalize these justifications for users. 
	%(see step 7 in Figure \ref{img_overview}). 
	% If at least one clause applies, we can restrict our attention to the positive clause's justifications.
	%\item We verbalize both the learned clauses from step 7 as well as any generated justifications from the previous step as natural language text.
	\item[$\Rightarrow$] We can now iteratively present an image with the corresponding mask, classification, and verbalizations to a user, who is prompted to accept or reject the classification as well as each justification thereof individually. Based on this user feedback, we augment our training data to update the current classification hypothesis.
\end{enumerate}

% We can now iteratively present a new image, its corresponding mask from step 3, its properties from step 5, the classification from step 8 and the verbalizations from the final step to a user. For each image, a user can then accept or reject the classification as well as each justification thereof individually. Based on this user feedback, the image is either simply added to the dataset with the corresponding label (in the ``wrong'' or ``right for the right reasons'' cases), or additional dummy data is generated that reflects the ``right for the wrong reasons'' case. We can then -- in the case of a wrong classification -- retrain from the newly augmented dataset and repeat.

\paragraph{} The resulting system combines the advantages of both knowledge-driven ILP and data-driven deep learning methods, and human expertise: The neural net component allows for extracting relevant information for a subsequent classification from image data; a format that is hardly amenable to alternative approaches. While training this component requires a large amount of data, this data only needs to be classified according to the presence of \emph{any} defects. Such data can usually be obtained relatively easily.
More detailed labelling regarding the precise locations and dimensions of individual defects is only necessary for a much lower number of data entries. Additionally, since the network is not used for the final classification, it only needs to be trained exactly once for any new domain of application.

Subsequently, the actual classification is provided by a logic program learned by the ILP component. Logic programs are intrinsically explainable, and a justification for a classification can be generated and presented in human readable form. The non-trivial data representation required for ILP can be generated automatically from the output of the neural component.

Finally, an ILP component can be retrained relatively inexpensively, allowing for integrating human expertise into the system by iteratively updating the ILP data set based on user feedback. In the following, we will describe these components in detail.

\section{The CNN-based Component for Defect Detection}\label{sec:cnn}

In the first component of the system, we aim to identify potential defects in an image of some given industrial component. The nature of the format of such data (i.e. image files) suggests training a convolutional neural network. We can safely assume that in any industrial setting, large amounts of image data of individual components can be easily obtained and classified as flawed or flawless, so that obtaining a sufficient amount of labelled training data for supervised learning is feasible.

Furthermore, a plurality of methods exist for localizing relevant parts of an image. Such methods fall broadly into one of two classes:

\begin{itemize}
\item \emph{Explainable AI (XAI) methods} include e.g. \emph{Lime} \cite{lime}, \emph{Layer-Wise Relevance Propagation} \cite{Montavon2019} and \emph{Grad-CAM} \cite{Selvaraju_2019}. These methods attempt to identify the parts of an image that are most relevant for the specific output of a (usually, but not exclusively) neural net classifier. 

Unfortunately, experiments with these methods on our data set have shown that the resulting image maps do not sufficiently reliably highlight the locations of actual defects.

\item \emph{Object Detection} models are instead trained directly to recognize, localize and/or classify specific objects in a given image input. These models are primarily used for e.g. identifying people in security camera footage or road signs in dashboard cameras, and have correspondingly received a lot of attention in recent years. Examples include the family of \emph{YOLO}-models \cite{redmon2016look}, variants of \emph{Mask RCNN} \cite{he2018mask} and the models included in the \emph{Tensorflow Object Detection API} \cite{huang2017speedaccuracy}. 

Unfortunately, these models often require large amounts of pixelwise-annotated training data, which is difficult and expensive to obtain. While few-shot approaches for object detection exist, these (even more than object detection models in general) suffer from the additional flaw, that they are pretrained on large amounts of highly \emph{heterogeneous} images, with the objects intended to be recognized being largely \emph{homogeneous}. In contrast, data sets such as ours are highly \emph{homogeneous}, whereas the features to be detected (i.e. flaws) are relatively \emph{heterogeneous}, making it substantially more difficult for object detection models to effectively learn to recognize them reliably from little training data.
\end{itemize}

\subparagraph{} This problem is also identified by \cite{Boi2021} regarding the same domain of application (i.e. defect detection in industrial components), and approached with a novel mixed-supervision model architecture, which trains on large amounts of binary classified training data (has defects / has no defects), combined with a small subset of pixelwise-annotated data. The resulting model outputs such a binary classification as well as an image mask highlighting the locations of defects in the image input.

\begin{figure}
\begin{center}
%\begin{tabular}{rl}
\includegraphics{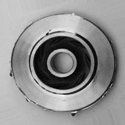} \qquad
\includegraphics{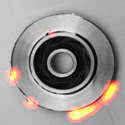}
%\end{tabular}
\end{center}
\caption{Example input image with its inferred mask.}
\label{img_ex}
\end{figure}

We consequently use their approach to train a new model on our data set for 20 epochs, augmented by manually annotated image masks for 86 of the training images. The resulting model is able to predict image masks on the 4211 training images with detectable flaws, allowing us to localize their present defects, see Fig.~\ref{img_ex}. Since this model is ultimately used for generating masks only, training statistics (which primarily relate to classification accuracy) are hardly informative, and hence omitted.

%\TODO{insert stats here? Do stats even matter, given that we don't care about the actual predictions?}

%Figures \ref{img_overview} and \ref{img_demo} 
%Fig.~\ref{img_ex} show an example of an input image with its inferred mask. 

\paragraph{Discussion} Which model architecture works best for a given use case depends on the makeup of the data set used. In \cite{limealeph}, it consists of simple geometric shapes, for which a vanilla convolutional network and Lime suffice. \cite{Boi2021} was specifically introduced for detecting defects in industrial components, so lends itself nicely to our use case, but might fail when the data set is less homogeneous, in which case object detection models will likely be more suitable.

Additionally, our model only highlights defects, but does not further classify them. Modifications of the model architecture used, or using object detection models instead, could boost the performance of the subsequent ILP component if in addition, the \emph{kinds} of defects can be identified by the statistical model already (e.g. holes, scratches, abrasions, etc.).

\section{The ILP-Based Component for Classification and Justification}

Our next goal is to extract relevant information about a given image datum from the image masks obtained from the neural net component, and represent this as a set of Prolog clauses for inductive logic programming.

To that end, we heuristically determine a suitable cut-off value (in our use case 0.3) below which we consider pixel values in an image mask (representing the model's certainty) to be equal to 0. We then determine the fully connected components (i.e. superpixels) of the resulting masks and compute various properties of both the image itself as well as the individual superpixels (each corresponding to a single defect), which we formalize in terms of Prolog predicates. 

For our use case, we decide on the following specifics:
\begin{itemize}
	\item \texttt{has\_hp}: a binary predicate associating a (name for an) image datum with its respective (names for) superpixel components.
	\item \texttt{has\_size}: a binary predicate associating a superpixel with the sum of its mask pixel values, representing its ``pixel$\cdot$certainty mass''.
	\item \texttt{distance\_from\_center}: a binary predicate representing the distance of a superpixel from the center of an image. This approximates the relevant aspects of the location (e.g. inner rim, outer rim) of a superpixel, assuming rotational symmetry and sufficient uniformity across the images (zoom, angle/skew etc.).
	\item \texttt{eccentricity}: a binary predicate associating a superpixel with $\sqrt{1-b^2/a^2}$, where \begin{enumerate*} \item $a$ is the maximum distance between two pixels $a_1,a_2$ in the superpixel and \item $b$ is the shortest distance between two pixels $b_1,b_2$ in the superpixel such that the line segment between $b_1$ and $b_2$ is orthogonal to the line segment between $a_1$ and $a_2$. \end{enumerate*}
	
	This approximates the ``roundness'' of a superpixel by treating it as an ellipsis, ideally allowing the ILP to distinguish between (round) holes and (less round) scratches or abrasions.
	\item \texttt{num\_hps}: a binary predicate associating an image datum with the total number of its superpixels.
	\item \texttt{total\_volume}: a binary predicate associating an image with the sum of the volumes of all of its superpixels.
\end{itemize}

Ideally, we would use the exact numerical values in all of the above predicates for ILP purposes. In practice, it is difficult for ILP systems to learn theories with numerical constants, even if (as in our use case) we can restrict their occurences to upper or lower bounds on variables, and to values occuring in the training data only. We therefore discretize the range of likely values for each of the above predicates using heuristically determined cut-off values and constants representing numerical intervals. For example, the values of the \texttt{has\_size}-predicate are let to be \texttt{vol\_small} (for values $<200$), \texttt{vol\_medium} (for values $200\leq n<900$) or \texttt{vol\_large} (otherwise). Additionally, we introduce a binary predicate \texttt{num\_leq} representing the $\leq$-relation between the thus introduced interval constants.

\iffalse
%num_hps(cast_6858,many_hps).

%has_hp(cast_6858,hp_6858_0).
%has_hp(cast_6858,hp_6858_1).

%has_hp(cast_6858,hp_6858_3).

%has_size(hp_6858_0,vol_small).
%distance_from_center(hp_6858_0,inner_rim).
%eccentricity(hp_6858_0,round_ish).

%has_size(hp_6858_1,vol_large).
%distance_from_center(hp_6858_1,outer_rim).
%eccentricity(hp_6858_1,round).

%has_size(hp_6858_3,vol_medium).
%distance_from_center(hp_6858_3,outer_rim).
%eccentricity(hp_6858_3,round).
\fi

\begin{figure}\vskip 4mm
\begin{lstlisting}[caption={Prolog representation of the image from Figure~\ref{img_ex}.},label={lst_prolog},basicstyle=\scriptsize,language=prolog]
total_volume(cast_6858,tvol_large).
has_hp(cast_6858,hp_6858_2).

has_size(hp_6858_2,vol_large).
distance_from_center(hp_6858_2,outer_rim).
eccentricity(hp_6858_2,very_elongated).

\end{lstlisting}
\end{figure}

Listing \ref{lst_prolog} shows an excerpt of the resulting Prolog representation of the example image used in Figure \ref{img_ex}, corresponding to steps 2 and 3 from Fig.~\ref{img_overview},  using the above predicates and interval constants for numerical values. It states that the image contains a total defective volume of size \texttt{tvol\_large}, and has a superpixel (i.e. defect) \texttt{hp\_6858\_2} (representing the abrasion at the outer rim) of volume \texttt{vol\_large}, being located at the \texttt{outer\_rim} and having an eccentricity of \texttt{very\_elongated}.

Finally, we classify a subset of our training data (in our case, the same 86 entries we manually annotated with image masks) more-or-less arbitrarily as ``ok'' or ``defective'' and use the ILP system \emph{Aleph} \cite{aleph} to induce an initial hypothesis (shown in Figure~\ref{img_overview} at step 4). In our experiments, we use a noise value of 10 and a search depth of 50 to obtain a theory consisting of three clauses and an accuracy of 98\% in less than ten seconds; a time frame well-suitable for iterative online retraining. Each returned clause represents a candidate for a sufficient condition for the targeted class (in our case: ``defective'').

%\begin{figure}
%\begin{lstlisting}[caption={An initial prolog theory learned by Aleph},label={lst_aleph},basicstyle=\scriptsize]
%true_class(A) :-
%  has_hps(A,B), distance_from_center(B,C), 
%  num_leq(C,inner_rim), eccentricity(B,D), 
% num_leq(D,round).
%true_class(A) :-
%  total_volume(A,B), num_leq(tvol_medium,B).
%true_class(A) :-
%  has_hp(A,B), has_size(B,C), 
%  num_leq(vol_medium,C), distance_from_center(B,D), 
%  num_leq(D,inner_rim).
%\end{lstlisting}
%\begin{center}
%\scriptsize``A component is considered \emph{defective}, iff \begin{enumerate*} \item it has at least one defect, which is located closer to the center than the inner rim and is at most round, or \item it has a total defective area of at least medium size, or \item it has at least one defect, which has at least medium size and is located closer to the center than the inner rim \end{enumerate*}''
%\end{center}
%\TODO{replace by actual verbalization}
%\end{figure}

\paragraph{Discussion} Naturally, which predicates we use strongly depends on the specifics of the use case. While we use Aleph, alternative systems might plausibly be better suited in situations where numerical values should occur as bounds in a proposed hypothesis. Alternatively, clustering algorithms can be used to determine the number and thresholds of interval classes, obviating the need to decide on them manually. We use \texttt{distance\_from\_center} and \texttt{eccentricity} as approximations of the relevant locations within the component in the image, and shapes of defects, respectively. In a real-life use case with less homogeneous data, classical image recognition methods are more appropriate for determining such properties, or, as mentioned in Section 1, can possibly be inferred by the preceding ML component already.

Notably, the inferred image masks and the manually annotated image masks differ significantly with respect to the properties used for Prolog predicates; most notably the pixel mass of superpixels: The manually annotated superpixels are precisely localized with sharp boundaries and values either 0 or 1, whereas the inferred superpixels are less precise, more spread out and have smoother boundaries. If we want to use both types of masks for training an ILP system, some scaling of the associated pixel values is correspondingly necessary to make the computed attribute values adequately consistent across both types.

Finally, there is a choice involved with respect to 
which classification is used as the \emph{positive} one explained by the learned theory: ILP systems attempt to learn a theory which entails as many \emph{positive} ($C^+$) examples and as few \emph{negative} ($C^-$) examples in the data set as possible. This matters insofar as the positive class is then explained as a set of Horn clauses, which together correspond to a \emph{disjunction of existentially quantified conjunctions of non-negated atoms}. The positive class should correspondingly be chosen as the one more likely explainable by such statements. We choose $C^+$=``defective'' as positive, since this classification is more likely the result of the \emph{existence} of a defect with certain properties.

%\TODO{Aleph vs Popper?}

\section{The Interactive Component}

Having obtained relevant properties of an input image and a first hypothesis for classification, our next goal is to provide a classification and a justification to a user in human-readable form, obtain feedback and integrate the latter back into the training process.

Naturally, Aleph outputs a Prolog theory, so we can use Prolog to obtain classifications based on the learned theory. In that case, however, we only obtain a boolean value. While we can still present a user with a verbalization of the learned theory as justification, it would be preferable to present a case-specific justification instead, instantiated for the exact image currently under consideration.

For that reason, we implement a custom method for evaluating the learned theory on a given image input which, besides evaluating each atom in a clause and instantiating variables with adequate values (e.g. the superpixels in an image), keeps track of whether an atom in a clause evaluates to \emph{true} or \emph{false} under a given variable assignment. While this is naturally less efficient than a dedicated logic programming language, the number of possible values for any given variable and the corresponding search tree are small enough that efficiency is not considerably relevant. As a result however, we obtain a concrete explanation (in the form of a list of true or false atoms) for a given image's classification, as shown in Figure \ref{img_overview} (step 6).

%\begin{figure}
%\begin{lstlisting}[caption={Classification and justifications for the image from Figure \ref{img_ex}},label={lst_justification},basicstyle=\scriptsize]
%Classification: true
%Justification 1: total_volume(cast_6858),tvol_large),
%  num_leq(tvol_medium,tvol_large)
%\end{lstlisting}
%\begin{center}
%\scriptsize``The component is classified as \emph{defective}, because it has a total defective area of at least medium size''
%\end{center}
%\TODO{replace by actual verbalization}
%\end{figure}
%
\begin{figure*}[htb]
\begin{center}
	\includegraphics{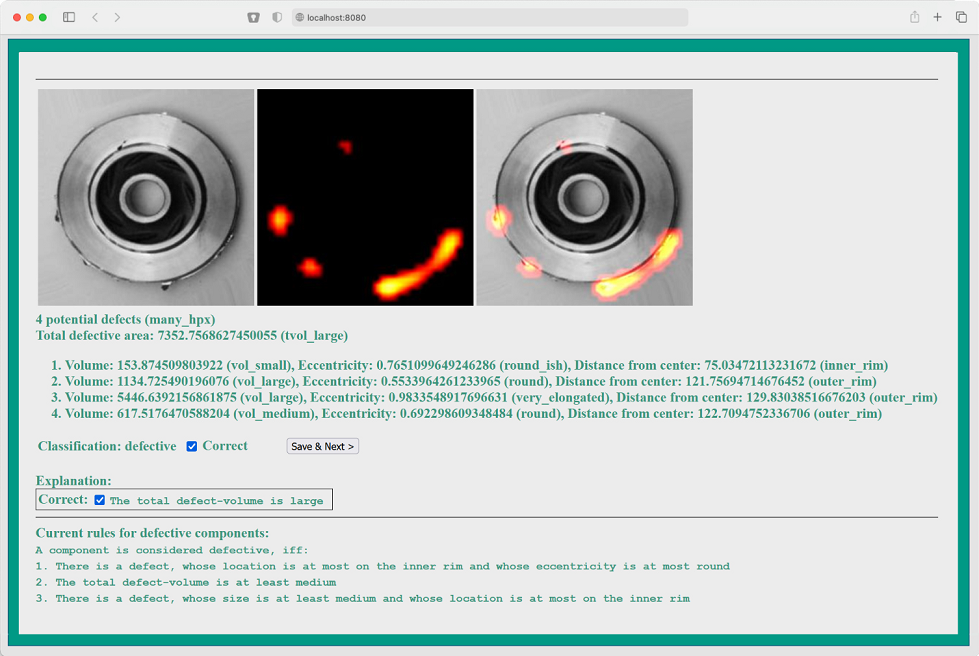}
\end{center}
	\caption{A screenshot of the demonstrator user interface.}\label{img_demo}
\end{figure*}

\subparagraph{} In order to obtain human-readable representations of both the current Prolog theory and the justifications, we need to associate each predicate with a natural language verbalization. This is largely straight-forward, except that we need to consider variable instantiations in \emph{functional} predicates jointly with subsequent atoms. More precisely: binary predicates that evaluate to \emph{true} if and only if the second argument is a uniquely determined value, such as \texttt{total\_volume} or \texttt{has\_size}, are representations of \emph{functions} in logic programming languages and (in our case) only occur in conjunction with subsequent predicate applications containing the same value, such as \texttt{total\_volume(A,B), num\_leq(B,tvol\_small)}. The corresponding ver\-ba\-li\-za\-tion therefore needs to adequately take into account, that these two atoms conjointly describe a property of \emph{the} (unique) total volume \emph{of} the element \texttt{A}. %(shown in Listings \ref{lst_aleph} and \ref{lst_justification}). 
Similar considerations apply to the \texttt{has\_hp}-predicate, which is used analogously to a non-deterministic function.

\subparagraph{} Finally, we need to integrate user feedback back into the system. This comes in the form of a boolean value for the classification itself as well as each justification. We distinguish three cases of user acceptance:
\begin{enumerate}
	\item Both classifications and justifications are accepted as correct; in that case we can simply add the image datum with its inferred classification to the training data.
	\item The classification is rejected. This necessarily entails that all justifications are false. We then add the image datum with the opposite label to the training data and retrain the ILP component to obtain an updated hypothesis.
	\item The classification is accepted, but (one or several) justifications are rejected (``right for the wrong reasons'').
\end{enumerate}
The last case deserves some closer attention:

%ILP systems attempt to find a theory that entails as many \emph{positive} examples (here: ``defective'') as possible and as few \emph{negative} examples (here: ``ok'') as possible. 
The ``right for the wrong reasons'' case means that an explanatory clause $E$ should not entail the associated classification $C^\pm$.
To ensure that, we can add a new dummy entry $D$, which should logically entail $\neg\forall x (E[x] \Rightarrow C^\pm[x])$ (or equivalently: $\exists x(E[x] \wedge \neg C^\pm[x])$) while remaining lo\-gic\-ally consistent with any alternative theory the ILP system might learn in the absence of $D$. Notably, ILP systems based on Horn clauses do not use \emph{negations} of atoms or whole clauses.

In the positive case $C^\pm=C^+$, the dummy entry should discourage Aleph from learning $E$ as a clause. It then suffices to have $D$ satisfy
$E$ and having ground truth label $C^-$, thus implying that Aleph incurs a penalty iff $E$ entails a clause of the learned theory. By not supplying any additional properties for $D$, adding $D$ does not otherwise impact Aleph's training procedure negatively.

In the negative case $C^\pm=C^-$ however, the only conclusion we can draw is that Aleph's learned theory is not \emph{complete} - which is to be expected, given a limited data set and arbitrarily complex \emph{true} theory. In that case, additional data is required to eventually increase the coverage of the learned theory.
%\TODO{Note: there's a choice involved which classification we choose as the \emph{positive} one (depending on whether Type I/II errors are more tolerable). Alternatively, one could have Aleph learn \emph{two} theories for each class; drawback: conflicting classifications are to be expected. $\Rightarrow$ Limitations/Future Work?}

\paragraph{Discussion} Some ILP systems allow for \emph{incremental learning} directly, improving a previously learned hypothesis based on new data alone without the need to reprocess the entire training set. While this clearly speeds up the retraining process, in our use case full batch training was sufficiently fast so that this aspect did not require further attention. This will likely become more relevant when scaling the system up to real-world applications.

For use cases where less computationally expensive (e.g. few-shot object detection) models are used, iterative updates to the statistical model can be included, for example via a mask editor. In the ideal case, the model would be updated immediately online, more realistically, it could be retrained periodically (e.g. nightly).

We mentioned in Section \ref{sec:cnn} that other candidate models for the CNN component can potentially be used to also classify the \emph{kinds} of defects (scratch, abrasion, etc.) occurring in some input image. Alternatively, we can have the ILP component learn predicates for these kinds of defects, which can be subsequently used in the ILP training data. While the final theory learned should be logically equivalent, using the new predicates in the training data would lead to more informative descriptions and justifications given to a user. Additionally, it would allow for more detailed user-feedback, in that the classifications of the individual defects (and the justifications thereof) could be accepted/rejected in the exact same manner as the final classification and their justifications.

\section{Demonstrator User Interface}

Figure \ref{img_demo} shows a screenshot of the user interface for our implementation. It uses a python-based web framework, serving a web page accessible via a browser. The original image and the image mask are shown both separately and overlayed, with verbalizations of the defects, the current theory (bottom of the screenshot), the classification and explanations for the classification shown. The classification and the explanations have check boxes that a user can uncheck to indicate their invalidity. Clicking on the ``Save \& Next''-button registers the user feedback, (optionally) updates the knowledge base and displays the data for the next image.

The code for our demonstrator and a link to a publicly running instance will be made available at \url{https://gitlab.cc-asp.fraunhofer.de/sees/iaai-cai-2021}.

\section{Conclusion and Future Work}

We have presented a proof-of-concept of an interactive AI-driven support system for industrial quality control, that extends visual defect recognition with verbal explanations by combining deep learning with inductive logic programming.
Besides the possible improvements and design choices discussed in the sections on the individual components, we plan to extend the approach to other use cases. Both performance and general usefulness of the system can be improved by including more background knowledge, possibly on the basis of more detailed ontologies. Additionally, the explanatory component can be extended by and combined with various other approaches, for example \emph{contrastive} and other example based explanations \cite{Rabold21}.
Finally, we plan to evaluate our approach in terms of a user study with respect to applicability and usefulness of explanations in the manufacturing domain.

%\TODO{Future work/limitations: numerical values for ILP, alternatively: clustering methods for boundaries between interval classes, proper image recognition methods for superpixel stuff, classify defects (hole, tear etc.) either in the NN component already or via image recognition algorithms, incremental learning vs batch retraining, positive class (see TODO 9) -- mabye also a separate section \emph{discussion/parameters/design choices}?}
%\TODO{``Resize your graphics before you include them with LaTeX. You may not use trim or clip options as part of your \textbackslash includegraphics command''}

\bibliography{biblio.bib}

\end{document}